\documentclass[sigconf, nonacm]{acmart}

\usepackage{graphicx}
\usepackage{amsmath}
\usepackage{amsthm}
\usepackage{booktabs}
\usepackage{algorithm}
\usepackage{algorithmic}
\usepackage[switch]{lineno}

\usepackage{mathrsfs}
\usepackage{xspace}
\usepackage{float}
\usepackage{color,xcolor}
\usepackage{enumitem}
\usepackage{subcaption}
\usepackage{bm}
\usepackage{stfloats}
\usepackage{multicol}

\newcommand{\up}{\textcolor{red}{$\bm{\uparrow}$}}
\newcommand{\down}{\textcolor{green}{$\bm{\downarrow}$}}
\newcommand{\same}{\textcolor{blue}{$\bm{-}$}}
\newcommand{\model}{C\lowercase{ohort}N\lowercase{et}\xspace}
\newcommand{\MFLMfull}{Multi-channel Feature Learning Module\xspace}
\newcommand{\MFLM}{MFLM\xspace}
\newcommand{\CDMfull}{Cohort Discovery Module\xspace}
\newcommand{\CDM}{CDM\xspace}
\newcommand{\CRLMfull}{Cohort Representation Learning Module\xspace}
\newcommand{\CRLM}{CRLM\xspace}
\newcommand{\CUMfull}{Cohort Exploitation Module\xspace}
\newcommand{\CUM}{CEM\xspace}

\newcommand{\rv}[1]{{\color[HTML]{000000}#1}}

\begin{document}

\title{
\model: Empowering Cohort Discovery for Interpretable Healthcare Analytics}

\author{Qingpeng Cai}
\affiliation{%
  \institution{National University of Singapore}
  \country{Singapore}
}
\email{qingpeng@comp.nus.edu.sg}

\author{Kaiping Zheng}
\affiliation{%
  \institution{National University of Singapore}
  \country{Singapore}
}
\email{kaiping@comp.nus.edu.sg}

\author{H. V. Jagadish}
\affiliation{%
  \institution{University of Michigan}
  \city{Michigan}
  \country{United States}
}
\email{jag@umich.edu}

\author{Beng Chin Ooi}
\affiliation{%
  \institution{National University of Singapore}
  \country{Singapore}
}
\email{ooibc@comp.nus.edu.sg}

\author{James Yip}
\affiliation{%
  \institution{National University Health System}
  \country{Singapore}
}
\email{james_yip@nuhs.edu.sg}


\begin{abstract}
Cohort studies are of significant importance in the field of healthcare analysis. However, existing methods typically involve manual, labor-intensive, and expert-driven pattern definitions or rely on simplistic clustering techniques that lack medical relevance. Automating cohort studies with interpretable patterns has great potential to facilitate healthcare analysis but remains an unmet need in prior research efforts. In this paper, we propose a cohort auto-discovery model, \model, for interpretable healthcare analysis, focusing on the effective identification, representation, and exploitation of cohorts characterized by medically meaningful patterns. \model initially learns fine-grained patient representations by separately processing each feature, considering both individual feature trends and feature interactions at each time step. Subsequently, it classifies each feature into distinct states and employs a heuristic cohort exploration strategy to effectively discover substantial cohorts with concrete patterns. For each identified cohort, it learns comprehensive cohort representations with credible evidence through associated patient retrieval. Ultimately, given a new patient, \model can leverage relevant cohorts with distinguished importance which can provide a more holistic understanding of the patient's conditions. Extensive experiments on three real-world datasets demonstrate that it consistently outperforms state-of-the-art approaches and offers interpretable insights from diverse perspectives in a top-down fashion.

\end{abstract}



\maketitle

\section{Introduction}
\label{sec:intro}

Effective patient analysis in healthcare~\cite{zhou2020clinical, huang2020cohort, wang2020clinical, yu2023predict, zheng2024exploiting} is pivotal for unlocking the vast potential inherent in electronic health records (EHRs). 
To facilitate clinicians in analyzing EHR data, it is crucial to explore cohorts that characterize patients sharing common conditions or responses to treatments. 
Studying the commonalities of cohorts among their associated patients enables clinicians to uncover valuable relationships and future outcomes. When assessing new patients, taking their relevant cohorts as auxiliary information empowers clinicians to gain a deeper understanding of their conditions, enhancing the quality of patient care.
The significance of cohorts with insights has been confirmed in a multitude of studies, underscoring the importance of incorporating cohort studies into healthcare analysis.

Cohort studies entail the identification of a specific group of patients, referred to as a ``\textbf{cohort},'' based on a discernible ``\textbf{pattern}''. Subsequently, this cohort is compared to other groups without such a pattern. 
Traditionally, these cohort-defining patterns are manually crafted by clinical experts, relying on features with specific value ranges (e.g., blood pressure $>$ 140/90 mmHg) and are associated with outcomes of interest, such as disease progression and endpoints.
Although these manual processes can yield medically meaningful insights, 
they demand extensive medical knowledge, and the subsequent identification of outcomes is both time- and resource-intensive.
As such, incorporating automated approaches~\cite{ooi2015singa} for the auto-discovery of cohorts characterized by meaningful patterns is imperative for healthcare analysis.

Unfortunately, cohort discovery within the context of EHR data presents several challenges that have not been effectively addressed in prior studies. 
Cohort identification calls for the creation of interpretable and generalizable medical patterns, which should be defined at a granular level, e.g., feature level. Existing studies employ various approaches to form cohorts, including  matching based on static features ~\cite{qin2021retrieval,ma2022retrieval, pahins2019coviz}, rule-based formulations~\cite{zhang2022rudi}, and analyses of event trajectories~\cite{omidvar2020cohort}.
However, these approaches are not effective for uncovering the underlying patterns in time-series features, as they fail to capture dynamic changes or feature interactions over time. Further, many features are recorded as numerical values, rendering the discovery of cohorts more challenging.
To handle such values, conventional approaches, such as discretization, not only require specific value range definitions but also constrain the search space for exploration, leading to suboptimal performance.

Some recent machine learning based approaches propose to
employ clustering~\cite{zhang2021grasp, panahiazar2015using, yu2023predict} or patient similarity~\cite{suo2018deep, zhu2016measuring} to group
patients into distinct ``cohorts''. However, these approaches predominantly rely on latent representations at a coarse level, typically the patient level, lacking concrete and interpretable patterns. The absence of interpretability restricts their practical utility for clinicians.
In some other studies~\cite{zhang2021grasp, omidvar2018cohort, omidvar2020cohort}, cohorts are created based on a small number of patients (e.g., a batch) for efficiency, which are not credible enough to convince clinicians.

\begin{figure}[t]
\centering
\includegraphics[width=0.49\textwidth]{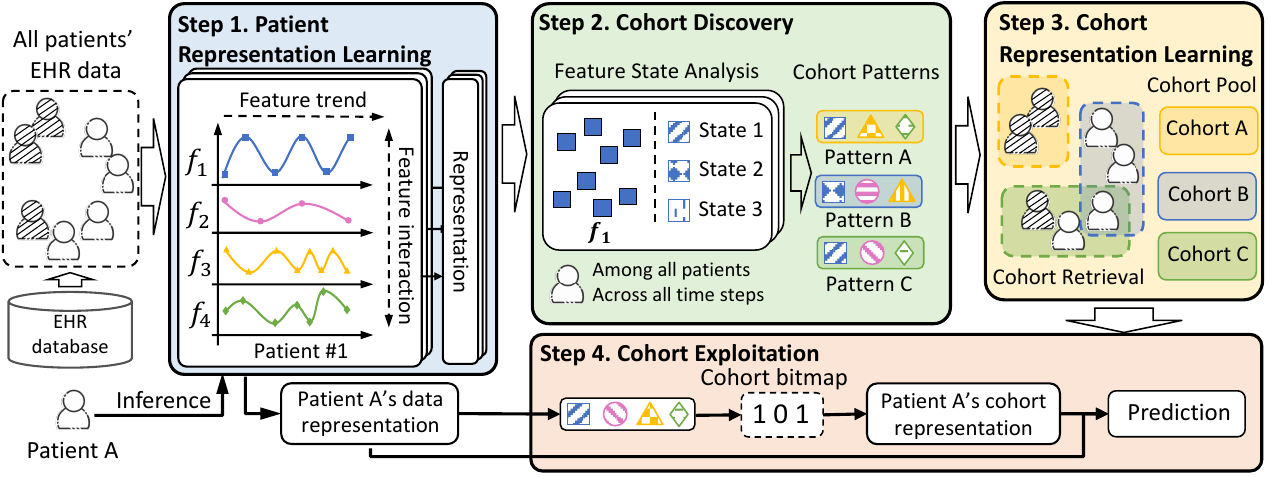}
\vspace{-5mm}
\caption{The systematic pipeline of cohort learning
}
\label{fig:pipeline}
\vspace{-5mm}
\end{figure}

To address above challenges, we propose an effective cohort discovery model, namely \model, with the following contributions.

\begin{itemize}[leftmargin=*]
    \item \model supports effective cohort discovery without the infusion of external knowledge. Specifically, it can identify, represent, and utilize medically significant cohorts with concrete feature patterns, enhancing prediction and interpretation simultaneously.
    To the best of our knowledge, cohort discovery with interpretable patterns, despite its importance, is not attainable in the past.

    \item \model features a systematic pipeline for healthcare analysis, centered around four innovative modules. Specifically, \MFLMfull initially learns feature-level representations through fine-grained modeling of feature trends and interactions, thereby facilitating subsequent feature state analysis.
    Through an exploration of the interrelationships among features, in \CDMfull, we devise a heuristic cohort exploration strategy for the auto-discovery of cohorts with concrete feature patterns. 
    For each pattern, \model retrieves its patients in \CRLMfull, forming its cohorts with comprehensive representations.
    Ultimately, by modeling patients' individual data and identifying their associated cohorts with distinguished importancethrough \CUMfull , \model can provide personalized assessment with interpretable insights.
    
    \item We evaluate its effectiveness on three real-world EHR datasets.
    Extensive experimental results validate that, by taking into account discovered valuable cohorts, it consistently outperforms state-of-the-art baselines with significant improvements.
    Meanwhile, when making decisions, it can offer medically interpretable insights from different perspectives that are consistent with findings 
    in previous studies.
    
\end{itemize}
In the remainder of this paper, we review related work in Section~\ref{sec:rw} and present the methodology of \model with its modules in Section~\ref{sec:framework}. Following this, we elaborate experimental results in Section~\ref{sec:exp} and conclude in Section~\ref{sec:conclusion}.

\section{Related Work}
\label{sec:rw}

\textbf{EHR Data Modeling.}
The burgeoning of deep learning and its extensive utilization in healthcare analysis~\cite{hinton2018deep} has spurred researchers to develop multifarious advanced deep learning models with libraries~\cite{ooi2015singa, paszke2019pytorch} for modeling EHR data. Specifically, recurrent neural networks (RNN) such as long short-term memory (LSTM)~\cite{hochreiter1997long} and gated recurrent units (GRU)~\cite{chung2014empirical} are widely adopted for handling time-series EHR data.
Several studies have designed specific mechanisms to improve EHR data modeling performance and interpretability. 
For instance, for the consideration of interpretation, RETAIN~\cite{choi2016retain} employs two GRUs in reverse time order to distinguish attention at the visit and variable levels, while Dipole~\cite{ma2017dipole} adopts attention mechanisms to differentiate relationships among patients' visits. Besides, ConCare~\cite{ma2020concare} embeds each medical feature separately and designs a multi-head self-attention mechanism to learn the interrelationships among medical features.
From another perspective, some approaches focus on capturing specific concepts in EHR for boosted performance.
T-LSTM~\cite{baytas2017patient} devises a time-aware LSTM to capture the irregular time intervals in EHRs, and StageNet~\cite{gao2020stagenet} integrates stage modules to extract and utilize the information about disease stages.
In contrast to above studies that model patients' representations solely from individual EHR data,
our model can also discover medically meaningful cohorts from EHR data, learning overall information, and leverage these learned cohorts to enhance performance and interpretability.

\vspace{1mm}
\noindent
\textbf{Clustering.}
Clustering methods aim to group similar samples within a dataset based on similarity metrics. Approaches such as K-Means clustering~\cite{macqueen1967some}, hierarchical clustering~\cite{johnson1967hierarchical}, and co-clustering
~\cite{dhillon2001co} have demonstrated satisfactory performance across diverse applications.
Unfortunately, their learned cluster results lack interpretable definitions, such as feature-based patterns, which hampers their acceptance by clinicians.
Besides, most clustering methods fall short when analyzing EHR data due to their limited consideration of temporal sequences. 
While some studies have attempted to incorporate temporal relationships~\cite{vandromme2020biclustering, cachucho2017biclustering, watanabe2015biclustering},
they struggle to adequately address the complexities inherent in multivariate time series data from diverse patients - a challenge we aim to tackle in processing EHR data. 
Although these methods may perform well within specific datasets, their applicability to assessing new samples is non-trivial.

A few studies have attempted to facilitate deep learning tasks with these learned clusters. For instance, GRASP~\cite{zhang2021grasp} uses the K-Means algorithm to group batches of patients' overall representations into several clusters and employs the K-Nearest neighbor (KNN) to utilize these clusters. PPN~\cite{yu2023predict} clusters all patients' representations and selects the patients closest to centroids as typical patients, potentially deviating from centroids. However, both learned clusters and typical patients are identified and utilized via similarity scores (e.g., Euclidean distance) measured based on patients' learned representations, which lack explicit feature patterns and fine-grained interpretability.
In contrast, we focus on discovering universal cohorts characterized by interpretable feature patterns.

\vspace{1mm}
\noindent
\textbf{Cohort Study.}
Cohort studies, as a cornerstone of observational research in medicine,
typically require experts to manually define concrete patterns in statistics or relevant medical features~\cite{pahins2019coviz, pastor2021looking} to determine cohorts and their constituent patient populations.
For example, authors in~\cite{wu2020elevation} use glucose to define groups and investigate their associated endpoints. 
However, crafting such explicit feature patterns could be time-consuming and labor-intensive. 

To automate cohort studies, in~\cite{omidvar2020cohort}, a heuristic sequence matching algorithm is proposed to derive cohort representation by aggregating patients' trajectories.
However, this approach can be intricate and computationally intensive when generating representative cohorts. Further, it is primarily suitable for analyzing patients' actions or events (such as disease occurrence or medication usage) and is not applicable to more complex numerical medical features.
Besides, some studies attempt to identify similar patient groups by exclusively relying on static characteristics, such as gender and occupation. For example, RIM~\cite{qin2021retrieval} and RB-GBDT~\cite{ma2022retrieval} match identical non-time series features in structured data to identify similar patients. RuDi~\cite{zhang2022rudi} categorizes patient groups by rules. It designs operators for features and converts knowledge from black-box models into rule-based student models. However, precise matching has restricted potential in learning cohorts for EHR data which comprises time-series medical features with numerical values.

Differing from these existing approaches, 
we propose \model that employs deep learning techniques to automatically identify, learn, and exploit cohorts, facilitating clinical decision-making processes with boosted performance and interpretable medical insights.

\section{Methodology}
\label{sec:framework}

\begin{figure}[t]
\centering
\includegraphics[width=0.45\textwidth]{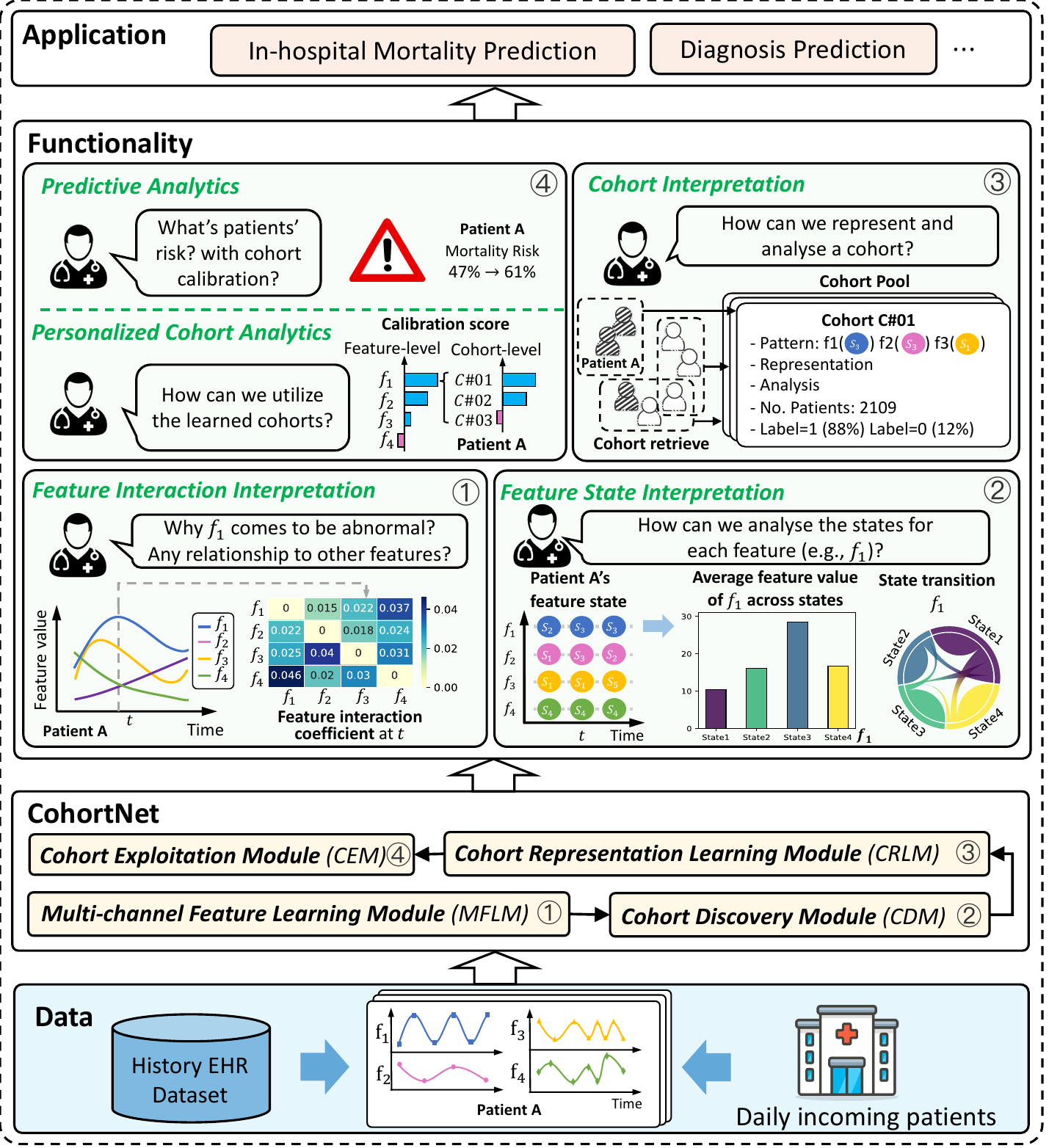}
\vspace{-3mm}
\caption{An overview of using \model for the auto-discovery of cohorts in healthcare analysis}

\label{fig:framework}
\vspace{-5mm}
\end{figure}

\subsection{Overview}
We use Patient A as a running example in Figure~\ref{fig:framework} to illustrate how to employ \model on EHR data to discover cohorts and make predictions, providing functionalities for medical interpretation. 

\noindent
\textbf{Data.} We initially collect comprehensive EHR data from hospital, including lab tests and vital signs, focusing on patients like Patient A. This dataset is pivotal for learning patients' data representations and discovering significant cohorts.

\noindent
\textbf{\model.} comprises four advanced modules:
(a) \textit{\MFLMfull} (\MFLM) 
utilizes patients' individual EHR data to learn informative representations by separately modeling each feature while simultaneously capturing feature interactions and trends;
(b) \textit{\CDMfull} (\CDM) 
first analyzes the states of features and then employs a heuristic cohort exploration strategy to unveil substantial cohorts characterized by specific feature patterns;
(c) \textit{\CRLMfull} (\CRLM) 
retrieves each cohort's associated patients by utilizing its pattern and further analyzes its representations while extracting other essential information (e.g., label distributions);
(d) \textit{\CUMfull} (\CUM) 
identifies a patient's relevant cohorts with differentiated importance and generates personalized cohort representations for calibration.
More details about \model are in Section~\ref{sec:methodology}.

\noindent
\textbf{Functionality.} 
Our \model emphasizes interpretability, demonstrating its ability to provide medically interpretable insights. By using Patient A as a case study, we illustrate several key functionalities. The \textit{Feature-level Interaction Interpretation} enables a nuanced understanding of Patient A's medical features and their interactions, essential for assessing abnormalities. For each feature, the \textit{Feature State Interpretation} offers a better understanding of its possible states, interpreted by distinct value ranges, transition pathways, and interactions. Our model acquires the ability to discover an extensive pool of cohorts, and each supported by credible evidence via \textit{Cohort Interpretation}. By identifying Patient A's relevant cohorts, we can further conduct \textit{Personalized Cohort Analytics}, enabling tailored analysis at both cohort and feature levels. This leads to more precise predictions in our \textit{Predictive Analytics}, leveraging these insights for enhanced medical outcomes. For more details on these interpretative functionalities, please refer to the Appendix.

\noindent
\textbf{Application.} Our model can tackle diverse downstream healthcare applications. 
In this paper, we focus on two extensively studied tasks, namely in-hospital mortality prediction and diagnosis prediction, to demonstrate how it facilitates clinicians in delivering accurate predictions with interpretability.

\vspace{-1mm}
\subsection{Preliminaries} 
\label{sec:methodology}
Before the introduction of \model, we summarize the notations in Table~\ref{tab:notation} and list the necessary definitions below.

\begin{table}[t]
  \centering
  \caption{Notations used in this paper.}
  \vspace{-3mm}
  \begin{tabular}{cl}
    \toprule
    Notation & Definition \\
    \midrule
    $\mathbf{X}$  &   EHR data \\
    $\mathbf{m}$           &   masking vector for medical features \\
    $\mathbf{e}$  &   feature embedding  \\
    $\mathbf{u}$ &  representation of feature interaction \\
    $\mathbf{v}$ &  representation of feature trend  \\
    $\bm{\alpha}$     &  attention scores of a feature's interactions \\
    $\mathbf{o}$ &  fused representation at the feature level\\ 
    \midrule
    $\mathbf{h}$        &  overall EHR data representation at the feature level \\ 
    $\mathbf{h}'$        &  overall cohort representation at the feature level \\ 
    $\Tilde{\mathbf{h}}$  &  overall EHR data representation at the patient level \\ 
    $\hat{\mathbf{h}}$    &  overall cohort representation at the patient level \\ 
    \midrule
    $s$             &  feature state \\
    $\bm{\psi}$     &  pattern mask \\
    $\bm{\eta}$     &  cohort pattern \\
    $\mathcal{C}(\bm{\eta})$     &  cohort's overall representation \\
    $\xi$               &  cohort \\
    $Pool(\xi)$     & cohort pool         \\
    $\bm{l}$        & labels 
    \\  
    $\bm{\beta}$    & attention scores of a feature's relevant cohorts\\
    $\mathbf{b}$    & personalized cohort bitmap \\
    $z$             & overall calibration score \\
  \bottomrule
\end{tabular}
\label{tab:notation}
\end{table}

\noindent
\textbf{EHR data.}
A patient's EHR data comprises 
multivariate time series
collected over time. Each time series denotes a medical feature that may have varying frequencies and lengths. 
We process each time series at regular intervals, resulting in each patient's data being represented as sequences of medical features, denoted as $\mathbf{X} = \{\mathbf{x}_1, \mathbf{x}_2, ..., \mathbf{x}_{|F|}\}$. $\mathbf{x}_i \in R^T$ represents the sequence of the $i$-th medical feature over $T$ time steps, and $|F|$ denotes the number of medical features.
Furthermore, we introduce a masking vector $\mathbf{m} \in \{0, 1\}^{|F|}$ where $m_i=0$ means that the $i$-th feature never presents in a patient's EHR data.
\rv{To lay the groundwork for our investigation, we first define several core concepts
as below.}

\vspace{-\topsep}
\begin{definition}[Cohort] 
A cohort is defined as a tuple ${\xi}=\langle\bm{\eta}, \mathcal{C}(\bm{\eta})\rangle$ where $\bm{\eta}$ denotes the cohort pattern utilized for patient retrieval, and $\mathcal{C}(\bm{\eta})$ signifies the cohort representation used in prediction tasks. 
A patient is classified into a cohort if and only if this patient manifests the specific cohort pattern at any time step.
\end{definition}
\vspace{-\topsep}

\vspace{-\topsep}
\begin{definition} [Cohort Pattern] \label{def:cohort-pattern}
Each cohort is characterized by a distinctive cohort pattern, comprising several medical features along with their respective states, formally defined as $\bm{\eta}=\{(i, s_i)\}^{d}$, where $i$ is the feature index, $s_i$ represents the state of the $i$-th feature, and $d$ indicates the number of involved features in the pattern.
\end{definition}
\vspace{-\topsep}

\vspace{-\topsep}
\begin{definition}[Cohort Representation] \label{def:cohort-rep}
Each cohort's representation, denoted as $\mathcal{C}(\bm{\eta})$, is derived from the retrieved patients' representations within the cohort and is subsequently utilized in predictive analyses of patient conditions. 
\end{definition}
\vspace{-\topsep}

To facilitate automated cohort identification, learning, and exploitation, we outline a predictive task comprising four sequential steps: patient representation learning, cohort discovery, cohort representation learning, and cohort exploitation, with details below.

\noindent\textbf{Patient Representation Learning.}
Given a patient's EHR data $\mathbf{X}$, this step to learn a mapping function $\mathcal{F}$to generate the patient's overall representation $\Tilde{\mathbf{h}}$ by summarizing feature-level representations $\mathbf{h}$.
It also strives to encapsulate essential patient information, including attention scores $\bm{\alpha}$ of feature interactions, derived from the patient's EHR data $\mathbf{X}$, i.e., $\Tilde{\mathbf{h}}, \bm{\alpha}=\mathcal{F}(\mathbf{X}, \mathbf{m})$.

\noindent\textbf{Cohort Discovery.}
For each feature, the step of cohort discovery initially utilizes a clustering algorithm to partition each feature into distinct states $s$.
Subsequently, a cohort exploration strategy is employed in function $\mathcal{G}$ to explore a cohort pool $Pool(\xi)$ considering interactions from learned attention scores $\bm{\alpha}$, where each cohort exhibits a specific pattern $\bm{\eta}$ as per Definition~\ref{def:cohort-pattern}.
This step is formalized as $Pool(\xi)=\mathcal{G}(\bm{\alpha}, \mathbf{h})$.

\noindent\textbf{Cohort Representation Learning.}
Upon identifying a cohort pattern $\bm{\eta}$, a representation learning algorithm $\mathcal{C}$ is applied to retrieve all associated patients and learn the corresponding cohort representations (as per Definition~\ref{def:cohort-rep}) from these patients.

\noindent\textbf{Cohort Exploitation.}
This step of cohort exploitation examines a patient's EHR data $\mathbf{X}$ alongside the cohort pool $Pool(\xi)$. It begins by identifying each patient's pertinent cohorts through a cohort bitmap $\mathbf{b}$, subsequently deriving the patient's personalized cohort representation $\hat{\mathbf{h}}$.
The final prediction $\Tilde{y}$, computed using the function $\mathcal{U}$, integrates the patient's individual data representation $\Tilde{\mathbf{h}}$ 
with the patient's cohort representation $\hat{\mathbf{h}}$ as calibration. This function is denoted as $\Tilde{y} = \mathcal{U}(\Tilde{\mathbf{h}}, \hat{\mathbf{h}})$.

Our overall objective is to learn a cohort pool $Pool(\xi)$ while concurrently learning the mapping functions $\mathcal{F}$, $\mathcal{G}$, $\mathcal{C}$, and $\mathcal{U}$ to accurately predict the outcome $\Tilde{y}$. To streamline the presentation, we frame our prediction task as a binary classification without loss of generality,
where $\Tilde{y} \in \{0,1\}$. To enhance readability, we omit the superscript $t$ when analyzing a single time step.

\subsection{\MFLMfull}
\begin{figure}[t]
\centering
\includegraphics[width=0.45\textwidth]{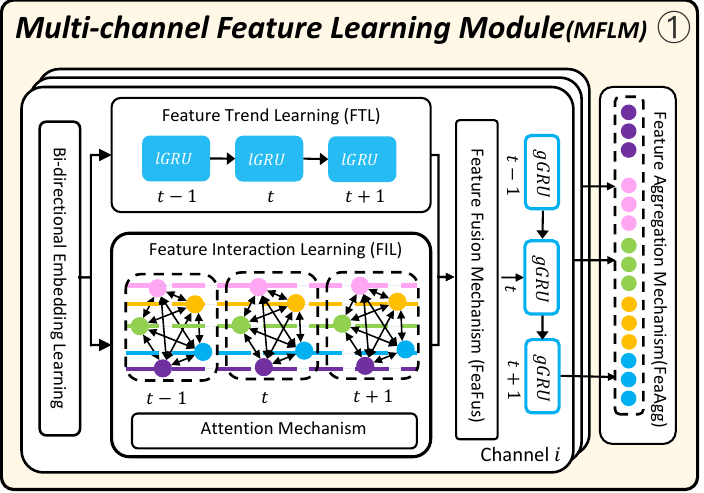}
\vspace{-2mm}
\caption{The \MFLMfull. 
}
\label{fig:MFLM}
\vspace{-4mm}
\end{figure}

Some prior methods for processing EHR data,  such as Channel-wise LSTM~\cite{Harutyunyan2019} and ConCare~\cite{ma2020concare}, 
conceptualize 
a patient as a composite of her/his medical features. 
They employ separate models to encode the raw values of each medical feature, thereby preserving the individuality of each feature. 
However, these raw values are insufficient for modeling 
complicated feature conditions or feature interactions~\cite{cai2022elda}.
In contrast to these methods, \MFLM, as depicted in Figure~\ref{fig:MFLM}, 
attains fine-grained representations of EHR data at the feature level, learning feature trends and interactions across time.

To preserve each feature's individuality, \MFLM employs multiple channels to learn a patient's overall representation, with each channel focusing on a particular feature.
In each channel, we first convert each medical feature's numerical values into meaningful embedding vectors $\mathbf{e}$ via a Bi-directional Embedding Learning (BiEL) mechanism~\cite{cai2022elda} that has been validated to be beneficial for embedding numerical features with a predefined bounds $a$ and $b$:
\begin{equation}
\mathbf{e}_i = \left\{
\begin{aligned}
\frac{1}{b-a}(\mathbf{V}^a_{i}(x_i'-a) + \mathbf{V}^b_{i}(b-x_i')) & , & \text{if } \mathbf{m}_i=1, \\
\mathbf{V}^m_{i} & , & \text{if } \mathbf{m}_i=0.
\end{aligned}
\right.
\end{equation}
where $\mathbf{V}^a, \mathbf{V}^b, \mathbf{V}^m \in \mathbb{R}^{|F|*d_e}$ are embedding matrices that convert $x_i'$ into a lower-dimensional feature representation, and $|F|$ and $d_e$ denote the number of features and dimension of $\mathbf{e}$.
This mechanism preserves the advantages of a linear embedding mechanism while simultaneously controlling the embedding scale and enhancing the embedding quality.
Besides, we also integrate the Feature Interaction Learning (FIL) mechanism~\cite{cai2022elda}
which enables the capture of explicit feature interactions ($\mathbf{u}$) and their corresponding attention scores ($\bm{\alpha}$) using fewer training parameters. Taking the $i$-th feature for illustration, the FIL mechanism models its interactions with all the other features at each time step:
\begin{equation}
\begin{split}
(\mathbf{u}_{i,1}, \alpha_{i,1}), (\mathbf{u}_{i,2}, \alpha_{i,2}), ..., (\mathbf{u}_{i,|F|}, \alpha_{i,|F|}) 
= FIL(\mathbf{e}_1, \mathbf{e}_2, 
..., \mathbf{e}_{|F|})
\end{split}
\label{equ:alpha}
\end{equation}
By learning these fine-grained feature interactions, \model can capture diverse valuable feature patterns, 
facilitating the subsequent cohort auto-discovery processes.
Further, the temporal trends of medical features also are essential in the analysis. To capture such trends, we propose a Feature Trend Learning (FTL) mechanism that utilizes separate local GRUs (lGRU) to learn the temporal dynamics of each feature.
Specifically, given $i$-th feature's embeddings,
FTL leverages an individual $lGRU_i$ to model its temporal behaviors $\mathbf{v}$:
\begin{equation}
\mathbf{v}_{i}^1, \mathbf{v}_{i}^2, ..., \mathbf{v}_{i}^t = lGRU_i(\mathbf{e}_i^1, \mathbf{e}_i^2, ..., \mathbf{e}_i^t), t=\{1,2,...,T\}
\end{equation}

Next, to derive the overall representation of a certain feature $\mathbf{o}$,
we further devise a Feature Fusion (FeaFus) mechanism in \MFLM to incorporate feature embedding $\mathbf{e}$, feature interactions $\mathbf{u}$, and feature trend $\mathbf{v}$ through a multilayer perception.
\begin{equation}
\mathbf{o}_i^t = FeaFus(\mathbf{e}_i^t, \mathbf{u}_i^t, \mathbf{v}_i^t)
\end{equation}
Through this adaptive fusion mechanism, we derive a fused representation $\mathbf{o}$ for each medical feature with reduced dimensionality, facilitating computations for the following cohort discovery.

Subsequently, to learn each feature's overall EHR representation $\mathbf{h}$, \MFLM employs separate global GRUs (gGRUs) in each channel to maintain the learning focus:
\begin{equation}
\mathbf{h}_i^1, \mathbf{h}_i^2, ..., \mathbf{h}_i^t = gGRU_i(\mathbf{o}_i^1, \mathbf{o}_i^2, ..., \mathbf{o}_i^t), t=\{1,2,...,T\}
\end{equation}
Finally, we propose a feature aggregation ($FeaAgg$) mechanism that integrates the representations of all medical features learned in their respective channels into a holistic representation $\Tilde{\mathbf{h}}^t$.
\begin{equation}
\Tilde{\mathbf{h}}^t = FeaAgg(\mathbf{h}_1^t, \mathbf{h}_2^t, ..., \mathbf{h}_{|F|}^t) 
\label{equ:patient_rep}
\end{equation}
In $FeaAgg$, we compress and concatenate all feature-level representations to form the overall patient-level representation, allowing for a fine-grained understanding of the patient's feature conditions.

In \model, \MFLM serves as a solid foundation, facilitating following cohort discovery. Besides, it supports fine-grained feature-level interaction interpretation, depicting diverse abnormal feature conditions.

\subsection{\CDMfull}~\label{sec:cohort-discovery}
\begin{figure}[t]
\centering
\includegraphics[width=0.45\textwidth]{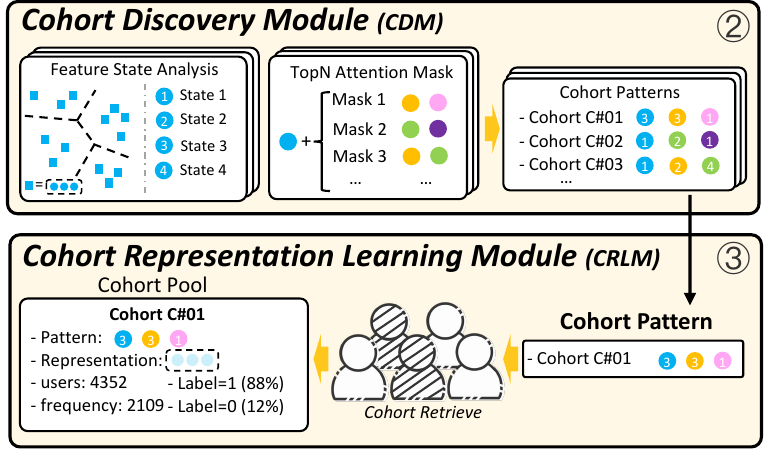}
\vspace{-2mm}
\caption{The \CDMfull and the \CRLMfull.
}
\vspace{-5mm}
\label{fig:CDM_CRLM}
\end{figure}

In medical studies~\cite{pahins2019coviz, pastor2021looking,wu2020elevation}, 
clinicians traditionally engage in manual pattern identification to classify patients into distinct cohorts, which is time-consuming and labor-intensive, requiring extensive domain knowledge. 
Further, in~\cite{omidvar2020cohort}, formalizing cohorts as trajectories of patient events, particularly in EHR data containing numerical values, poses significant challenges. Beside, \cite{xie2020cool} porposes an efficient cohort analysis engine, but without domain knowledge, it cannot discover meaningful cohorts. As for learning algorithms, while K-Means clustering employed in~\cite{zhang2021grasp, yu2023predict} can group patients based on their learned overall representations, the centroids learned in clusters are limited in interpretability.
To overcome these constraints, we introduce \CDM to automate the identification of universal cohorts exhibiting concrete feature patterns across multiple sequences of EHR data.
The overview of this module is shown in Figure~\ref{fig:CDM_CRLM}.

On account of the necessity of leveraging specific features in patterns, \CDM begins by classifying each feature into distinct states based on the learned representations $\mathbf{o}$. 
To achieve this, for each feature, \CDM gathers all its possible representations from all samples at all time steps and then employs a clustering algorithm, denoted as $Cluster$, to analyze and identify the feature into distinct states.
\begin{equation}
    \{s_0, s_1, s_2, ..., s_k\} = Cluster(\mathbf{o}_i, m_i, k) \label{equ:cluster}
\end{equation}
Compared with alternative clustering techniques such as hierarchical clustering and co-clustering, we ultimately select K-Means in this module due to its superior efficiency, and the centroids learned in K-Means are easier to apply when assessing new patients. 
In detail, we employ the K-Means algorithm in an adaptive manner to identify $k$ distinct states $s_i$ for the $i$-th medical feature. 
Different states of the same medical feature reflect distinct conditions, which can be approximately characterized by the average feature values of each state, feature interactions, and state transition pathways. 
As part of our algorithm, we particularly classify missing features (i.e., $m_i=0$) into a distinct state $s_0$ since the absence of a feature directly represents a particular condition.

Relying solely on a feature's state to represent the medical pattern can be inaccurate, while analyzing all medical features introduces the curse of dimensionality. More specifically, when extracting $n'$ medical features from EHR data and clustering them into $k$ clusters for each feature, the search space expands exponentially to $\mathcal{O}(k^{n'})$, leading to a increased computational complexity and a higher risk of overfitting in pattern discovery.

In practical medical scenarios, when analyzing the feature abnormality, clinicians jointly consider several other highly related features along with their values to achieve a more accurate assessment. Inspired by it, we address the cohort pattern exploration challenges by introducing a heuristic cohort exploration strategy in \CDM that can leverage feature interactions as guidance without the infusion of external medical knowledge. Specifically, it identify substantial cohort patterns by analyzing both feature interactions (i.e., $\bm{\alpha}$ learned in \MFLM) and their corresponding feature states.

When processing $i$-th medical feature, \CDM analyzes its feature interactions by transforming attention scores $\bm{\alpha}$ into an attention-based pattern mask $\bm{\psi}_i = \{0, 1\}^{|F|}$:
\begin{equation}
    \bm{\psi}_i = topN(\bm{\alpha}_i, n) + onehot(i) \label{equ:topN}
\end{equation}
where $topN(\cdot, \cdot)$ generates a binary vector identifying $n$ relevant medical features with the highest attention scores.
Additionally, $onehot(\cdot)$ constructs another binary vector where only the $i$-th dimension is set to one.
In \CDM, each medical pattern $\bm{\eta}$ is derived from the combination of both the attention-based mask $\bm{\psi}$ and the feature states $\mathbf{s}$, $\bm{\eta}_i = \mathbf{s} \odot \bm{\psi}_i$
where $\odot$ denotes the element-wise product. With such designed cohort exploration strategy employed in \CDM, \model can effectively find meaningful medical patterns without covering the whole search space, and each pattern $\bm{\eta}$ involves $d_{\psi}$ features, $||\bm{\psi}_i||_1 = d_{\psi} = n+1$.

\subsection{\CRLMfull}
\label{subsec:CRLM}
In prior studies, cohorts are delineated through various means, encompassing patient event trajectories~\cite{omidvar2020cohort}, hand-crafted medical definitions~\cite{pahins2019coviz, pastor2021looking,wu2020elevation}, and K-Means centroids~\cite{zhang2021grasp}. 
However, these approaches either lack compatibility with deep learning techniques
or encounter challenges regarding interpretability. In contrast, \model not only discerns cohorts with medically interpretable feature patterns but also furnishes informative cohort representations for predictive tasks.
Specifically, derived from its pattern $\bm{\eta}_i^q$ (the $q$-th pattern for the $i$-th feature), the \CRLMfull (as shown in the bottom of Figure~\ref{fig:CDM_CRLM}) retrieves the patients who exhibit this particular pattern and learns representations from these patients' commonalities.

In \model, a patient belongs to the cohort $\xi_i^q$ if and only if all states of the involved features in its pattern $\bm{\eta}_i^q$ exactly match the patient's feature states at a certain time step. 
With the retrieved patients, we derive the cohort's latent representation $\mathcal{C}(\bm{\eta}_i^q)$ by learning from these patients' representations and labels:
\begin{equation}
    \mathcal{C}(\bm{\eta}_i^q) = [\frac{1}{|\xi_i^q|} \sum_p h_i^p;\ \bm{l}_i^q],\ p\in \xi_i^q \label{equ:cohort_representation}
\end{equation}
where $p$ is a specific patient in the cohort $\xi_i^q$, and $[\cdot;\cdot]$ represents a concatenation function; $\bm{l}_i^q$ refers to the distribution of patients' task-relevant labels (e.g., mortality labels and diagnosis labels) and task-irrelevant labels (e.g., frequencies and demographic distributions) within this cohort.
Hence, a cohort in \model comprises both the feature-based concrete medical pattern and the general representation, denoted as $\xi_i^q = \langle\bm{\eta}_i^q, \mathcal{C}(\bm{\eta}_i^q)\rangle$.
Such an in-depth understanding of cohorts enables our model to deliver interpretable insights when leveraging these cohorts for personalized analysis. 
With numerous patterns discovered in \CDM, our \CRLM will create a substantial cohort pool, denoted as $Pool(\xi) = \{ \xi_i^1, \xi_i^2,...,\xi_i^{q}, ... \} \in \mathbb{R}^{|F|*|C|}$ where $|C|=|C_1|+|C_2|+...+|C_{|F|}|$, and $|C_i|$ denotes the number of cohorts discovered for $i$-th medical feature.

Additionally, we integrate several filters to narrow down the search space and mitigate the computation complexity, e.g., the sample frequency filter, to identify and exclude medical patterns that occur infrequently in the training samples, since low frequencies result in insufficient evidence to support these cohorts' credibility.

\subsection{\CUMfull}
\rv{The acquired cohorts should be readily applicable to new patients efficiently and effectively. However, the cohorts learned in earlier work~\cite{zhang2021grasp} have been limited by their applicability to specific patient populations.
Furthermore, cohorts in~\cite{omidvar2020cohort} require substantial computational resources due to reliance on sequence matching methods.
In \model, 
\CUM
can efficiently leverage these credible and universal cohorts as supplementary information when assessing individual patients' feature conditions.}
That is, it takes the cohort pool $Pool(\xi)$ as extensive knowledge and indexes the patient's relevant cohort with differentiated importance.
As a result, \model not only improves the performance of downstream prediction tasks but also
provides clinicians with interpretable cohort-based insights.

\begin{figure}[t]
\centering
\includegraphics[width=0.45\textwidth]{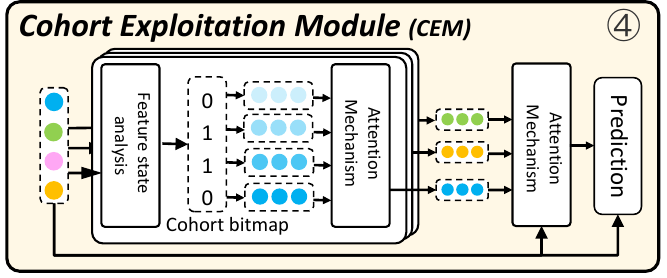}
\vspace{-2mm}
\caption{The \CUMfull. 
}
\vspace{-5mm}
\label{fig:CUM}
\end{figure}

As shown in Figure~\ref{fig:CUM}, \CDM first examines patients' feature states $\mathbf{s}$ at each time step and subsequently analyze cohort bitmap $\mathbf{b}_i = \{0,1\}^{|C_i|}$ when analyzing $i$-th medical feature: 
\begin{equation}
    \mathbf{b}_i^q = 1,\ if \ \exists \ \mathbf{s}_i^t \odot \bm{\psi}_i^q = \bm{\eta}_i^q,\ t=1,2,...,T  \label{equ:cohort_identification}
\end{equation}
where $\bm{\psi}_i^q$ is the pattern mask indicating the involved features in the pattern $\bm{\eta}_i^q$.
When integrating these cohorts to facilitate the evaluation of the $i$-th feature, it is crucial to distinguish the importance (denoted as $\bm{\beta}_i^q$) of the involved cohorts. To address this, we devise an attention mechanism that discerns the varying significance of different cohorts in relation to the target prediction task:
\begin{equation}
    {\beta'}_i^q = (\mathbf{W}_Q \cdot \mathbf{h}_i^T)\cdot( \mathbf{W}_K \cdot \mathcal{C}(\bm{\eta}_i^q)), \ q = 1,2,...,|C_i|
\end{equation}
\begin{equation}
    \beta_i^1, \beta_i^2, ..., \beta_i^{|C_i|} = softmax({\beta'}_i^1, {\beta'}_i^2, ..., {\beta'}_i^{|C_i|})
\end{equation}
\begin{equation}
    \mathbf{h}'_i = \sum_q \beta_i^q (\mathbf{W}_V \cdot \rv{\mathcal{C}(\bm{\eta}_i^q)})
\end{equation}
where $\mathbf{W}_Q$, $\mathbf{W}_K$, and $\mathbf{W}_V$ are trainable weights, \rv{and $\mathcal{C}(\bm{\eta}_i^q)$ denotes the cohort representation derived in Equation~\ref{equ:cohort_representation}}.
\CUM then summarizes a feature's cohort representations by weighing them with the learned attention scores $\bm{\beta}$, to derive the feature's cohort representation $\hat{\mathbf{h}}_i$.
Similar to \MFLM,
we concatenate all features' cohort representations and generate the patient's overall cohort-related representation $\hat{\mathbf{h}}=[\mathbf{h}'_1;\mathbf{h}'_2;...;\mathbf{h}'_{|F|}]$.

Finally, based on a patient's overall data representation $\Tilde{\mathbf{h}}$ learned in \MFLM and the patient's individual overall cohort representation $\hat{\mathbf{h}}$ learned in \CUM, we conduct a binary prediction task via:
\begin{equation}
    \Tilde{y} = \sigma(\mathbf{w}^p \cdot \Tilde{\mathbf{h}} + b^p + \mathbf{w}^c \cdot \hat{\mathbf{h}}) \label{equ:prediction}
\end{equation}
where $\mathbf{w}^p$, $b^p$, $\mathbf{w}^c$ are trainable weights, and $\sigma(\cdot)$ is the sigmoid function. 
When deriving the final prediction via Equation~\ref{equ:prediction}, we can re-express $\mathbf{w}^c \cdot \hat{\mathbf{h}}$ as a cohort-related overall calibration score $z$:
\begin{eqnarray}
    z &=&\mathbf{w}^c \cdot \hat{\mathbf{h}}  \\
     ~&=&\sum_i \underbrace{\mathbf{w}^c_i \cdot \mathbf{h}'_i}_{feature-level\ calibration\ score} \label{equ:f_cali} \\
     ~&=&\sum_i \sum_q \underbrace{ \mathbf{w}^c_i \cdot \beta_i^q (\mathbf{W}_V \cdot \rv{\mathcal{C}(\bm{\eta}_i^q)})}_{cohort-level\ calibration\ score} \label{equ:c_cali}
\end{eqnarray}
where the overall calibration score $\bm{z}$ can be decomposed into feature-level calibration scores, which can further be subdivided into cohort-level calibration scores.

Our model is capable of supporting diverse tasks, and we exemplify it by the binary classification task, where the binary cross-entropy loss function is employed:
\begin{equation}
    \mathcal{L} = -\frac{1}{B} \sum_{j=1}^B y_j log(\Tilde{y}_j) +  (1-y_j) log(1-\Tilde{y}_j)
\end{equation}
$y_j$ and $\Tilde{y}_j$ represent the ground truth labels and prediction labels, respectively, and B denotes the batch size. 

\section{Experiments}
\label{sec:exp}

\begin{figure*}[t]
\centering
\includegraphics[width=0.8\textwidth]{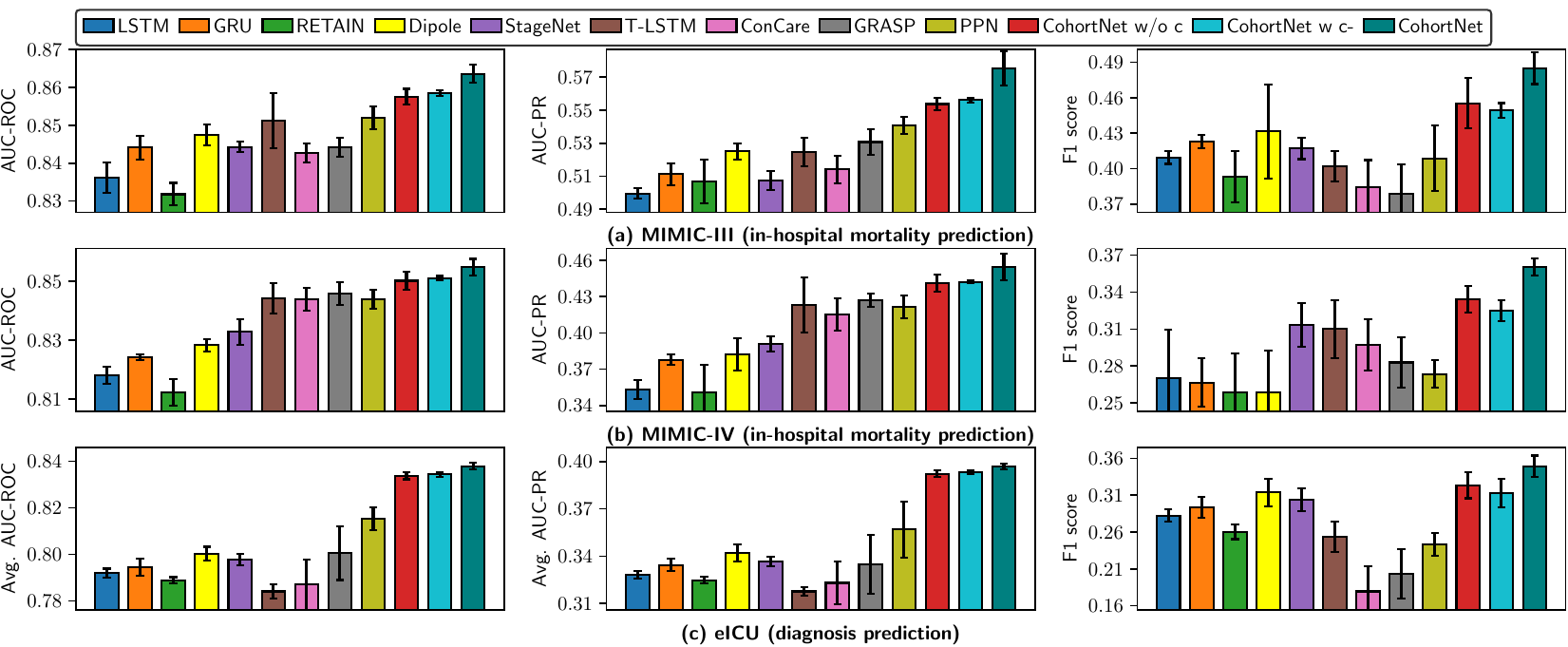}
\vspace{-2mm}

\caption{
\rv{Experimental results for in-hospital mortality prediction on the MIMIC-III and MIMIC-IV datasets and diagnosis prediction on the eICU dataset in terms of the three metrics. }
}
\vspace{-5mm}
\label{fig:mortality-results}
\end{figure*}

\subsection{Experimental Setup}
We evaluate the effectiveness of our model on three real-world benchmark EHR datasets: MIMIC-III, MIMIC-IV, and eICU. We perform the in-hospital mortality prediction, formulated as a binary classification task, on the first two datasets and conduct diagnosis prediction as a multi-label classification task on the third dataset.

\textbf{MIMIC-III Dataset}~\cite{johnson2016mimic} is a publicly available ICU dataset collected at Beth Israel Deaconess Medical Center spanning from 2001 to 2012. We first sample 21,139 admissions from this dataset based on patients' demographic data and EHR data as suggested by~\cite{Harutyunyan2019} and subsequently extract 63 clinically aggregated time-series medical features. related to vital signs and lab tests.

\textbf{MIMIC-IV Dataset}~\cite{johnson2023mimic}, a public contemporary EHR dataset released in 2023, provides valuable insights into a decade of admissions ranging from 2008 to 2019. Distinguished from MIMIC-III, this dataset comprises more recent data and a larger sample size. 
In particular, we extract 35,122 admissions from this dataset, 
with a focus on the top 70 most frequently occurring time-series features.

\textbf{eICU Dataset}~\cite{pollard2018eicu} is collected from many critical care units across the contiguous United States, encompassing patients admitted to these units during 2014 and 2015. In our experiments, we sample 41,547 admissions with 25 diagnosis labels as suggested by~\cite{sheikhalishahi2019benchmarking}, where each sample encompasses the 67 most frequently occurring features, which include 48 lab tests and 19 chart events.

In these datasets, each sample contains medical features recorded 
during the initial 48 hours following admission to the intensive care unit (ICU), 
and all features are applied a mean-std standardization.

\textbf{Evaluation Metrics.}
We compare \model with baseline models using the following metrics: the area under the receiver operator characteristic curve (AUC-ROC), the area under the precision-recall curve (AUC-PR), and the F1-score.
Among these metrics, AUC-PR is the primary metric because it is the most informative score when handling a highly imbalanced dataset~\cite{choi2018mime,zhang2021grasp,davis2006relationship}.
We divide the samples into $80\%$:$10\%$:$10\%$ for training, validation, and testing.

\textbf{Implementation Details.} Our experiments are conducted on a server with 48 Intel(R) Xeon(R) Silver 4214R CPU @ 2.40GHz and a GeForce RTX 3090 GPU.
For model training, we employ the Adam optimizer~\cite{kingma2014adam} and set the learning rate to 1e-3.

\textbf{Baselines.}  We compare \model with the baselines below.
\begin{itemize}[leftmargin=*]
    \item \textbf{LSTM}~\cite{hochreiter1997long} is an RNN-based model for time-series analysis.
    \item \textbf{GRU}~\cite{chung2014empirical} is another RNN-based time-series model but requires fewer parameters than LSTM.
    \item \textbf{RETAIN}~\cite{choi2016retain} utilizes two levels of GRU in the reverse time order
    to differentiate the importance of visits and variables.
    \item \textbf{Dipole}~\cite{ma2017dipole} adopts a bidirectional GRU and devises attention mechanisms to calculate the relationships among time steps. 
    \item \textbf{StageNet}~\cite{gao2020stagenet} models disease progression stages and incorporates them into learning disease progression patterns.
    \item \textbf{T-LSTM}~\cite{baytas2017patient} designs a time decay mechanism to handle irregular time intervals in EHRs.
    \item \textbf{ConCare}~\cite{ma2020concare} embeds each time-series medical feature separately and employs a self-attention model to learn the relationships among these features. 
    \item \textbf{GRASP}~\cite{zhang2021grasp} relies on a backbone model to learn patients' general representations, uses K-Means to find a group of similar patients, and applies K-NN to integrate the groups' information.
    \item \rv{\textbf{PPN}~\cite{yu2023predict} identifies typical patients to serve as prototypes and leverages these prototypes by calculating similarity metrics when assessing new patients.}
\end{itemize}
Moreover, we conduct an ablation study to assess the influence of our learned cohorts
by comparing it with the following baselines:
\begin{itemize}[leftmargin=*]
    \item \textbf{\model w/o c}
    removes all cohort-related modules, namely \CDM, \CRLM, and \CUM.
    This evaluation aims to validate the impact of the learned cohorts on the overall performance.
    \item \textbf{\model w c-} employs K-Means to cluster patients' overall representations directly in \CDM and then identifies patients' relevant cohorts directly via K-Means in \CUM. 
    The comparison demonstrates the necessity of defining cohorts with concrete patterns at the feature level.
\end{itemize}

\section{Results} \label{subsec:main-results}
We design experiments to answer the following questions:
\begin{itemize}
    \item \textbf{Q1: Analysis of model performance.} Can \model provide more accurate performance with cohort discovery? How sensitive is \model to core parameters?
    \item \textbf{Q2: Analysis of model interpretability.} How does \model provide functionalities for medical interpretation?
    \item \textbf{Q3: Analysis of model efficiency.} How about the efficiency of our \model?
\end{itemize}

\subsection{Q1: Analysis of model performance}
\noindent\textbf{Main Results.} The experimental results of \model compared to baseline models on the three datasets are illustrated in Figure~\ref{fig:mortality-results}.
Among the baselines, RETAIN exhibits inferior performance, as it prioritizes interpretability over performance. 
Dipole, StageNet, T-LSTM, and ConCare generally outperform the standard time-series models LSTM and GRU due to their elaborate mechanisms tailored for capturing specific characteristics in EHRs. 
\rv{GRASP and PPN enhance predictive performance by integrating representations of similar or typical patients.
In comparison to the best-performing baseline models,
\model consistently achieves improved performance in all metrics,
demonstrating substantial improvements in AUC-PR, with increases of 3.5\%, 2.8\%, and 4.1\% on the MIMIC-III, MIMIC-IV, and eICU datasets, respectively.} Furthermore, \model also achieves higher F1 scores by over 5\% on the in-hospital mortality prediction task.
The superior performance of \model stems from its capability of fine-grained feature representation learning and effectiveness of cohort discovery, learning, and exploitation.

\noindent\textbf{Ablation Study.} ~\label{subsec:ablation-study}
We further conduct an ablation study to compare the performance of \model and its several variants. As shown in Figure~\ref{fig:mortality-results},
\model w/o c, which includes only \MFLM, surpasses the other baselines, confirming the effectiveness of \MFLM in fine-grained feature-level representation learning by integrating feature interactions and feature trends. 
The comparison against this variant underscores the efficacy of \model, particularly in discovering significant cohorts defined by concrete feature patterns in \CDM, further learned in \CRLM, and ultimately exploited in \CUM. This comprehensive pipeline in \model leads to more accurate predictions, as indicated by improved F1 scores.
Furthermore, \model w c- demonstrates merely marginal improvement, indicating that directly clustering patients' overall representations cannot capture sufficient information for these prediction tasks. In contrast, \model achieves substantial improvements, emphasizing the necessity of fine-grained cohort learning at the feature level.

\begin{figure}[t]
\centering
\includegraphics[width=0.35\textwidth]{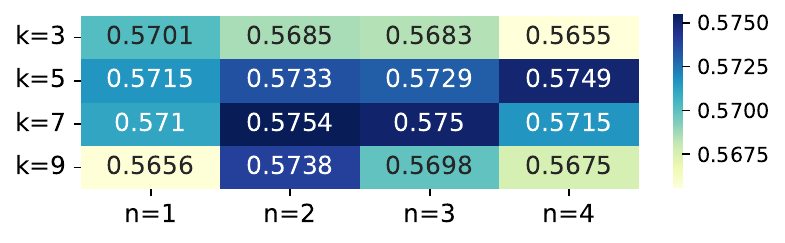}
\vspace{-2mm}
\caption{Sensitivity analysis of \model on $k$ and $n$ in terms of AUC-PR on the MIMIC-III dataset.
}
\vspace{-4mm}
\label{fig:sensitivity_AUCPR}
\end{figure}

\begin{figure}[t]
\centering
\includegraphics[width=0.5\textwidth]{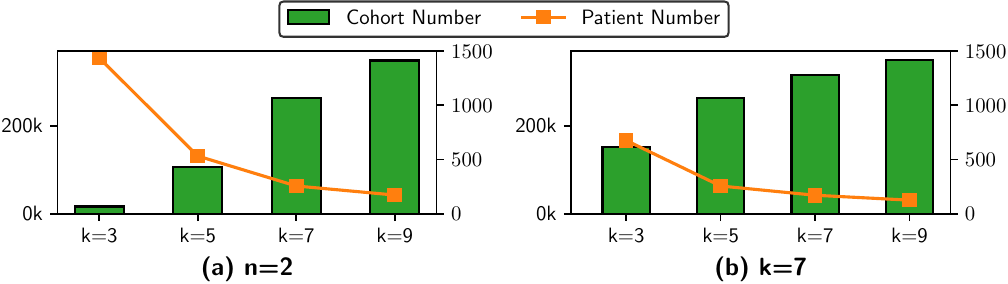}
\vspace{-5mm}
\caption{The cohort numbers and the average patient number in each cohort on the MIMIC-III dataset.
}
\vspace{-5mm}
\label{fig:sensitivity_kn}
\end{figure}

\noindent\textbf{Sensitivity Analysis.}~\label{sec:sensitivity_study}
In this experiment, we investigate the impact of two critical hyperparameters, $k$ in Eq.~\ref{equ:cluster} and $n$ in Eq.~\ref{equ:topN}, on the performance of \model as illustrated in Figure~\ref{fig:sensitivity_AUCPR}. 

Notably, \model consistently exhibits improvements across different settings compared with the best-performing baselines, which affirms the effectiveness of learning cohorts.
The highest AUC-PR value is achieved when $k=7$ and $n=2$.
We also observe that moderate values of $n$ and $k$ are crucial for achieving substantial performance improvements. 
This is because both hyperparameters influence the trade-off between cohort granularity and the patient number associated with each cohort.
As shown in Figure~\ref{fig:sensitivity_kn}, smaller values of $k$ or $n$ result in more general cohorts with larger patient numbers. However, such cohorts possess coarser-grained states and fewer selected features, thus preserving less personalized information. Conversely, larger values of $k$ or $n$ result in cohorts that maintain more fine-grained details and tend to attain improved performance. Nonetheless, selecting overly high values of $k$ or $n$ 
may result in overfitting, and consequently, degrade performance.

\begin{table}[t]
\caption{Statistics of cohorts w.r.t RR. 
}
\vspace{-3mm}
\setlength\tabcolsep{2pt}
\centering
\small
\begin{tabular}{ccccc}
    \toprule  
    Cohort &  Frequency & Patients & Pos-Rate & Cohort Pattern \\ 
    \midrule  
    C\#01 &  472    &  125     &  36.8\%   &  RR(S3\up); BUN(S2\up); PCO2(S7\up) \\
    C\#02 &  12519       &    2188     & 29.3\%   &  RR(S3\up); ALT(S3\down); AST(S2\down) \\
    C\#03 &  2909      &  1019       & 16.0\%   &  RR(S3\up); HCO3(S3\up); PCO2(S7\up) \\
    C\#04 &  35753      &   12858      & 12.1\%  &   RR(S2\same); HR(S7\same); HCO3(S2\same) \\
    \bottomrule  
\end{tabular}
\label{tab:rr_cohort}
\end{table}

\begin{figure}[tp]
\centering
\includegraphics[width=0.48\textwidth]{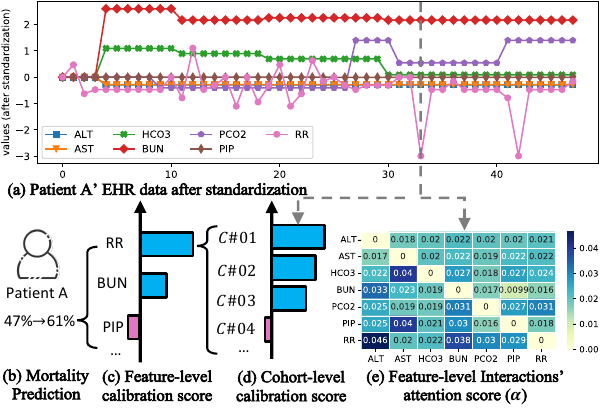}
\vspace{-5mm}
\caption{An example of how our model analyzes Patient A with interpretable insights at different levels.
Involved features include the respiratory rate (RR), alkaline phosphate (ALT), aspartate aminotransferase (AST), bicarbonate (HCO3), blood urea nitrogen (BUN), partial pressure of carbon dioxide (PCO2), peak inspiratory pressure (PIP). 
}
\vspace{-4mm}
\label{fig:personalize_cohorts}
\end{figure}

\subsection{Q2: Analysis of model interpretability}\label{subsec:interpretation}
In this section, we elucidate how our model interprets predictions with learned cohorts, offering medically significant insights from different perspectives in a top-down fashion, as elaborated in Section~\ref{sec:framework}. 
Figure~\ref{fig:personalize_cohorts}(a) showcases Patient A's standardized data for several critical features, with more details in the caption.

\vspace{1mm}
\noindent\textbf{Predictive Analytics.}
We start our exploration of its predictive analytics by examining Patient A. Notably, we observe that our proposed \MFLMfull predicts Patient A's mortality probability to be $47\%$, based solely on Patient A's EHR data representation $\Tilde{\mathbf{h}}$.
However, as shown in Figure~\ref{fig:personalize_cohorts}(b), upon identifying and leveraging Patient A's relevant cohorts, our entire model increases this predicted probability to $61\%$.

\vspace{1mm}
\noindent\textbf{Personalized Cohort Analytics.}
The changes in final predicted probabilities result from the examination of feature-level calibration scores, which are calculated in Eq.\ref{equ:f_cali}, representing the influence of each feature on the final prediction.
We illustrate these scores in Figure~\ref{fig:personalize_cohorts}(c), with positive scores depicted in blue and negative scores in pink.
Specifically, RR and BUN exhibit relatively higher feature-level calibration scores, thereby contributing to positive effects, while some other features (e.g., PIP) exert negative influences.

Each feature's feature-level calibration score is aggregated from the cohort-level calibration scores of its associated cohorts, as derived in Eq.~\ref{equ:c_cali}.
When analyzing the feature RR in Patient A, it identifies several relevant cohorts by evaluating features' states at different time steps. For instance, Cohort C\#01 is identified in the 34th hour via Eq.~\ref{equ:cohort_identification}.
As shown in Figure~\ref{fig:personalize_cohorts}(d), delving into the feature RR, we discover some crucial cohorts with different cohort-level calibration scores. Certain cohorts (e.g., Cohorts C\#01, C\#02, and C\#03) contribute to a higher risk of mortality, whereas others (e.g., Cohort C\#04) might not exhibit this tendency.
The details of all four relevant cohorts are presented in Table~\ref{tab:rr_cohort}, including the number of associated patients (Patients) and the rate of patients with positive labels, i.e., patients' mortality rate in this cohort (Pos-Rate). 

\vspace{1mm}
\noindent\textbf{Cohort Interpretation.} \label{subsec:cohort-interpretation}
Backed by the \CDMfull, each identified cohort has a distinct pattern comprised of multiple features along with their corresponding feature states.
For instance, Cohort C\#03 is characterized by the pattern where RR is in S3 (with lower values\down), HCO3 is in S3 (with higher values\up), and PCO2 is in S7 \up. More details about the feature states will be introduced in Section~\ref{subsubsec:feature state interpretation}.
This cohort is crucial for analyzing the reason behind Patient A's lower RR level and deteriorating health condition, as patients with such a pattern may be experiencing respiratory acidosis. Respiratory acidosis~\cite{epstein2001respiratory, dorman1954renal} is a condition characterized by an elevation in PCO2 in the blood.
Typically, it occurs when there is an accumulation of carbon dioxide in the body due to inadequate breathing or impaired gas exchange in the lungs. The low respiratory rate observed in the pattern suggests that patients are not breathing sufficiently to eliminate carbon dioxide effectively, resulting in elevated PCO2 levels. In response, the body compensates by increasing the concentration of HCO3 to maintain acid-base balance. 
In a nutshell, this cohort's characteristics align with medical knowledge surrounding respiratory acidosis, which further validates its interpretability in learned cohorts.

Considering the complexity of a patient's health condition, a patient may be associated with multiple cohorts that depict disease progression. 
As shown in Figure~\ref{fig:personalize_cohorts}(d),
Patient A is also associated with Cohort C\#02 and C\#01, which holds greater importance than C\#03.
Among them, Cohort C\#01 holds Patient A's highest cohort-level calibration score.
Despite representing only 125 patients, this cohort exhibits a significantly higher mortality rate (36.8\%) compared to other cohorts. Its cohort pattern features elevated BUN in S2 \up, indicating increased blood nitrogen from urea, which is regulated by the liver and excreted by the kidneys. Respiratory acidosis, affecting kidney function, may consequently alter BUN levels.
Existing medical studies~\cite{dorman1954renal,barker1957renal} explore the renal response to respiratory acidosis by analyzing relevant features like sodium and chloride. Given this context, it is promising to investigate this cohort, as abnormal features in its pattern may reveal possible interrelationships and inspire potential medical research advancements.

\model also identifies several common cohorts such as Cohort C\#04, which encompasses almost two-thirds of the patients in our training dataset. All features in its cohort pattern remain at a normal level (\same), reflecting the patients' normal conditions.
Hence, C\#04 demonstrates a slight negative influence on the final prediction when compared to the other cohorts with abnormal patterns.
By jointly considering multiple relevant cohorts with varied importance, \model can effectively facilitate clinicians with better assessments of the patient's condition.

\begin{figure}[t]
\centering
\includegraphics[width=0.48\textwidth]{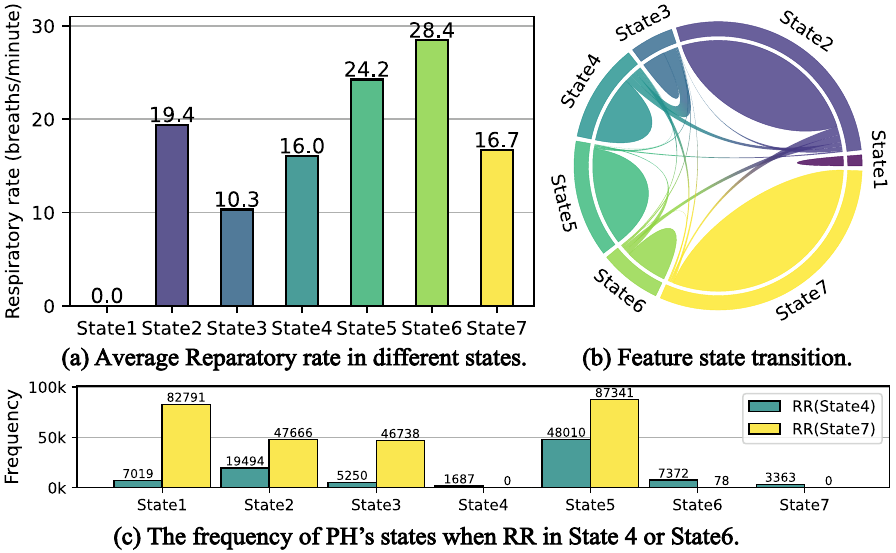}
\vspace{-5mm}
\caption{Feature state study in RR.
}
\vspace{-7mm}
\label{fig:feature_state}
\end{figure}

\vspace{1mm}
\noindent\textbf{Feature State Interpretation.}
\label{subsubsec:feature state interpretation}
In \CDMfull, each feature is classified into $k$ distinct states, and we take RR as an example to interpret feature states from three perspectives.

Firstly, the states of features generally correspond to varied medical values in EHRs. To account for this, we calculate the average values of RR across all patients in a state-wise manner, as illustrated in Figure~\ref{fig:feature_state}(a).
Among these states, there is a specific state, namely S1, which is assigned to patients who have not undergone testing or recording for RR.
Besides, we observe that different states of RR typically indicate different value ranges. For example, when a patient's RR significantly exceeds the upper limit of the normal range (20 breaths per minute), it tends to fall into S6, indicating the patient's relatively severe conditions.

Secondly, the transitions between the states of RR follow specific pathways, as shown in Figure~\ref{fig:feature_state}(b).
This diagram, based on all training samples, uses varying curve thicknesses to indicate the frequency of each state transition.
It is evident that not all state pairs have direct transitions.
For instance, there are one-way transitions from S2 to S3 and no direct transitions between S5 and S7.
These validate our capability to effectively capture the evolution of RR.

Thirdly, the differentiation between a feature's different states can be visualized by considering their co-existence with other features' states. For instance, as shown in Figure~\ref{fig:feature_state}(c), although S4 and S7 of RR exhibit similar average values, RR in S4 coexists with PH in S4 or S7, whereas RR in S7 never exhibit such coexistence.
Delving into state correlations, \model can capture more state distinctions with interpretable insights.

\vspace{1mm}
\noindent\textbf{Feature-level Interaction Interpretation.}
In addition to learning from the above cohorts, \model can capture diverse feature-level interactions through the \MFLMfull.
It determines the importance of these interactions, as measured by $\bm{\alpha}$ in Eq.~\ref{equ:alpha}. 
Figure~\ref{fig:personalize_cohorts}(e) illustrates an example where our model analyzes the feature RR, 
assigning higher significance to features with extremely abnormal values (such as BUN). Additionally, several other features (such as PCO2 and PIP) also receive elevated attention values, indicating their strong relevance to RR in the medical context. 
By integrating these crucial feature-level interactions, the representation learned in \model for each feature can depict the feature's various abnormal conditions, leading to improved performance and interpretability.

\subsection{Q3: Analysis of model efficiency}
\label{sec:efficiency}

\begin{figure}[tp]
\centering
\includegraphics[width=0.48\textwidth]{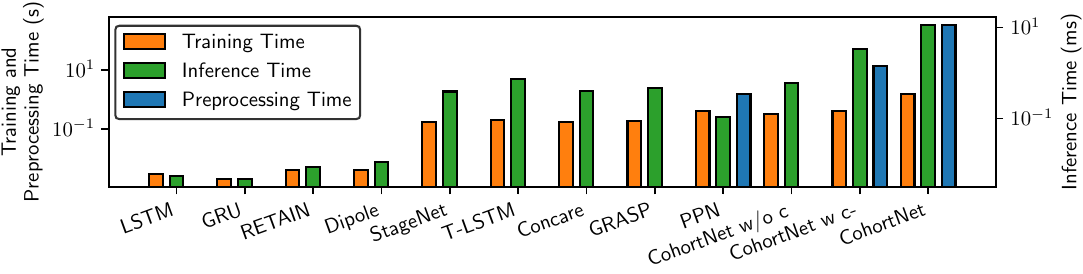}
\vspace{-5mm}
\caption{
Runtime of models on the MIMIC-III dataset.
}
\vspace{-5mm}
\label{fig:efficiency-compare}
\end{figure}

\noindent \textbf{Efficiency comparison.} 
We subsequently investigate the runtime of \model in comparison with the baselines on the MIMIC-III dataset. 
The runtime metrics include the training time required for a batch of patients, the inference time for a new patient, and the preprocessing time required in Steps 2 and 3 for training. The comparative runtime results are depicted in Figure~\ref{fig:efficiency-compare}.

Among the baseline models, GRU and LSTM stand out for efficiently handling time series data, demonstrating superior performance in training and inference times. However, more complex models such as RETAIN and Dipole, which feature dual-layer or bidirectional GRU architectures, introduce additional computational overhead. Likewise, specialized designs like StageNet and T-LSTM, which are tailored to model disease stage progression and irregular time intervals respectively, necessitate increased runtime.
Besides, compared with ConCare which also processes features separately, \model w/o c exhibits relatively prolonged runtime due to its consideration of fine-grained feature interactions and trends. 
Regarding cohort modeling, GRASP learns exclusively from batches of patients, whereas PPN, \model w c-, and \model entail additional preprocessing time to learn typical patients or cohorts. 

\begin{figure}[tp]
\centering
\includegraphics[width=0.48\textwidth]{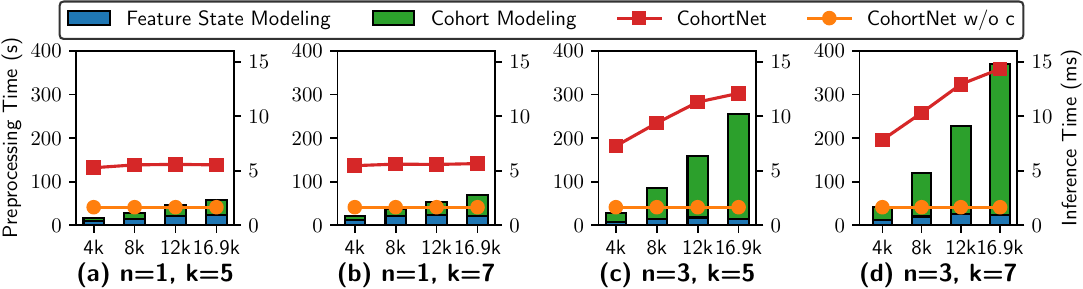}
\vspace{-5mm}
\caption{
Efficiency analysis on $k$ and $n$ in terms of runtime on the MIMIC-III dataset with different sample sizes.
}
\vspace{-5mm}
\label{fig:efficiency-study}
\end{figure}

\noindent \textbf{Effect of $k$ and $n$ on efficiency.}  We further evaluate the effect of $k$ and $n$ on our model's preprocessing efficiency for training and inference.
During training, it encompasses all four steps, with $k$ and $n$ specifically influencing in the preprocesses of feature state learning and cohort representations learning in Steps 2 and 3. Figure~\ref{fig:efficiency-study} shows that as the number of patients increases,
\model demands more time since it needs to process all patients.
Notably, when classifying each feature into fewer states (i.e., $k$=5) and 
exploring cohort patterns with less involved features (i.e., $n$=1),
training time does not drastically increase with a larger sample size.
It is attributed to the reduced cohort exploration spaces (i.e., lower $k$ and $n$). Beyond a certain patient count, 
the growth of discovered cohorts is gradual, indicating that the identified cohort patterns are capable of encompassing future patients.
However, \model with a larger $k$ needs a longer time to model features with more states.
Similarly, a higher $n$ indicates that more features will be considered when \model discovers cohort patterns. 
Consequently, as the training samples increase up to 16.9k, with a rise in both $n$ and $k$, \model discovers more cohorts, thereby elongating the time for subsequent patient retrieval and cohort learning.

During inference, altering sample size does not affect the time of \model w/o c, which relies solely on individual patient data.
In contrast, \model takes longer for predicting since it leverages discovered cohorts when making decisions.

\section{Conclusions}
\label{sec:conclusion}
In this paper, we present an interpretable healthcare analytics model, \model,
that facilitates
effective cohort discovery with concrete feature patterns, which is a crucial task not achieved in prior studies.
The model comprises four modules designed for fine-grained patient representation learning considering individual feature trends and feature interactions, the auto-discovery of cohorts with medically interpretable patterns, comprehensive cohort representation learning through associated patient retrieval, and personalized cohort exploitation for enhanced healthcare analysis.
Extensive experiments on three real-world EHR datasets show its superior performance, with AUC-PR improvements ranging from 2.8\% to 4.1\% compared to state-of-the-art baselines.
These results confirm \model's 
effectiveness in cohort discovery, learning, and exploitation for enhancing prediction performance and providing interpretable insights.
In future, we shall further collaborate with clinicians to validate our findings following medical practices.


\clearpage
\section{Appendix}
\appendix
\section{Interpretable Functionalities}
By prioritizing interpretability, we demonstrate our \model's capability to yield medically interpretable insights in Figure~\ref{fig:framework}.

\begin{itemize}[leftmargin=*]
    \item \textit{Feature-level Interaction Interpretation.} \rv{When assessing Patient A's abnormal medical features}, such as $f_1$, it is crucial to take into account its related features (e.g., $f_2$, $f_3$, $f_4$), allowing for a more accurate judgment of Patient A's conditions (e.g., any complications). Supported by devised \MFLM, \model can explicitly model these interactions across time with differentiated importance, facilitating 
    the analysis for each feature.

    \item \textit{Feature State Interpretation.} In \model, each medical feature will be grouped into distinct states through \CDM to facilitate subsequent cohort discovery and exploitation processes.
    In Figure~\ref{fig:framework}, each of Patient A's features, such as $f_1$, is classified into several distinct states, with each state inherently encapsulating unique meanings. For instance, states $s_1$, $s_2$, and $s_3$ of $f_1$ are delineated by varying value ranges, transition pathways, and interactions, 
    gaining a deeper understanding of the feature's conditions.
    Hence, these feature states establish a robust interpretability foundation for cohort creation and subsequent steps.
    
    \item \textit{Cohort Interpretation.} Through \CDM, our model acquires the ability to discover an extensive \rv{cohort pool}, where each cohort is characterized by a concrete pattern. In particular, each pattern consists of several medical features and their corresponding states. As an example depicted in Figure~\ref{fig:framework}, the pattern for Cohort C\#01 involves $f_1$ in $s_3$, $f_2$ in $s_3$, $f_3$ in $s_1$.
    For each cohort, it deepens its understanding through further exploration of its associated patients. Hence, all cohorts learned in our model are convincing with credible evidence.
    
    \item \textit{Personalized Cohort Analysis.} As \CRLM effectively learns significant cohorts, it is crucial to utilize these cohorts as references when providing personalized analysis.
    For instance, in the analysis of feature $f_1$ for Patient A, \CUM identifies relevant cohorts (e.g., C\#01, C\#02, C\#03, etc) tailored to Patient A's changing conditions. Analyzing these cohorts to examine the calibration scores at both cohort and feature levels, \CUM can offer a granular understanding of Patient A, facilitating more accurate predictions.
    
    \item \textit{Predictive Analysis.} 
    Our model analyzes patient health conditions using quantified prediction scores to provide timely alerts for potential deterioration. By leveraging Patient A's individual EHR data as the foundation (with a mortality risk of 47\%) and calibrating with relevant cohorts (increasing risk to 61\%), our model can effectively and accurately predict outcomes, offering medically interpretable insights.

\end{itemize}

\section{Model Complexity Analysis} \label{subsec:complexity}
In this section, we first analyze the complexity of each module and then explain the overall model complexity.

\textbf{\MFLMfull.} In this module, we compute the feature embedding in $\mathbf{e}$ in $\mathcal{O}(|F| \times T \times d_e)$, the feature trend $\mathbf{v}$ in $\mathcal{O}(|F| \times T \times (3d_ed_t+3{d_t}^2))$, the feature interaction $\mathbf{u}$ in $\mathcal{O}(|F|^2 \times d_e)$, the fused representation $\mathbf{o}$ in $\mathcal{O}(|F| \times (2d_e+d_t)d_o)$, the overall feature representation $\mathbf{h}$ in  $\mathcal{O}(|F| \times T(3d_o d_h+ 3{d_h}^2))$, and the overall patient representation $\Tilde{\mathbf{h}}$ in $\mathcal{O}(|F| \times (d_o \times d_p))$ where $d_e$, $d_t$, $d_o$, $d_h$, and $d_p$ denote the dimensions of $\mathbf{e}$, $\mathbf{v}$, $\mathbf{o}$, $\mathbf{h}$, and $\Tilde{\mathbf{h}}$, respectively. Overall, we achieve a complexity of $\mathcal{O}(|F| \times T \times (d_o d_h+ {d_h}^2))$ in this step.

\textbf{\CDMfull.} We cluster feature states in $\mathcal{O}(|F| \times (|P| \times T) k d_h I)$ and analyze patterns in $\mathcal{O}|F| \times (|P| \times T)\times d_{\psi} log((|P| \times T)\times d_{\psi}))$, where $|P|$ denotes the number of patients used for discovering cohorts. $k$ and $I$ refer to the number of clusters and iterations used in K-Means. 
Thus, the complexity of this module is $\mathcal{O}|F| \times (|P| \times T)\times d_{\psi} log((|P| \times T)\times d_{\psi}))$.

\textbf{\CRLMfull}. We learn comprehensive cohort representation in this module, and the complexity of this module is $\mathcal{O}(|F| \times (|P| \times T) \times (d_h+|L|) \times |C|)$ where $|L|$ and $|C|$ denote the size of used labels and learned cohorts.

\textbf{\CUMfull.} This module is to utilize patients' identified cohorts for final predictions, and the complexity of cohort exploitation in this module is $\mathcal{O}(|F| \times T\times|C|\times(d_h+|L|) + |F|\times(2d_h+|L|)) \sim \mathcal{O}(|F| \times T\times|C|\times(d_h+|L|)$.


Overall, We train our model with all the four aforementioned steps: learning patients' fine-grained representations via \MFLM in $\mathcal{O}(|F| \times T \times (d_o d_h+ {d_h}^2))$, discovering diverse cohort patterns via \CDM in $\mathcal{O}|F| \times (|P| \times T)\times d_{\psi} log((|P| \times T)\times d_{\psi}))$, representing cohorts via \CRLM in $\mathcal{O}(|F| \times (|P| \times T) \times (d_h+|L|) \times |C|)$, and utilizing these cohorts via \CUM in $\mathcal{O}(|F| \times T\times|C|\times(d_h+|L|))$. Therefore, the overall computational cost in the training phase is $\mathcal{O}(|F| \times (|P| \times T) \times (d_h+|L|) \times |C|)$.
However, when making inferences on new patients, only the first and last steps are required, leading to an overall complexity of $\mathcal{O}(|F| \times T\times|C|\times(d_h+|L|))$.

\section{Experimental Results}
\subsection{Scalability Analysis.}
~\label{sec:scalability}
\begin{figure}[tp]
\centering
\includegraphics[width=0.48\textwidth]{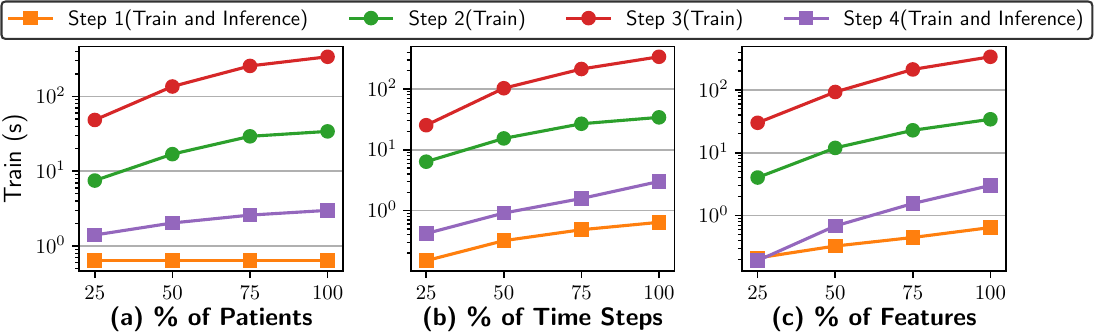}
\caption{
Scalability study of \model on varying numbers of patients, time steps, features on the eICU dataset.
}
\label{fig:scabability}
\end{figure}

As detailed in Section~\ref{subsec:complexity}, the time complexity of \model is influenced by the number of patients, time steps, and features. To evaluate its scalability, we vary these factors in the largest eICU dataset and measure the runtime of each step, as shown in Figure~\ref{fig:scabability}.

In Step 1, \model processes features across time steps to learn patients' overall representations, with computation time increasing linearly with the number of features and time steps.
In Steps 2 and 3, \model utilizes the complete data across all patients and time steps to identify cohort patterns and learn cohort representations, respectively. As illustrated in Figure~\ref{fig:scabability} (a) and (b), the runtime escalates more than linearly with an increase in the number of patients and time steps. This is because, with more patients and more time steps taken into consideration, \model identifies more cohort patterns and requires a longer time to derive the representations of these cohorts. Additionally, an increase in the number of features leads to the discovery of more potential feature interactions during the cohort modeling process, subsequently extending the runtime for these two steps, as depicted in Figure~\ref{fig:scabability} (c).
Meanwhile, as \model identifies an increased number of cohorts through these two steps, the computational demands also rise for training datasets or evaluating new patients in Step 4.

\begin{figure}[tp]
\centering
\includegraphics[width=0.5\textwidth]{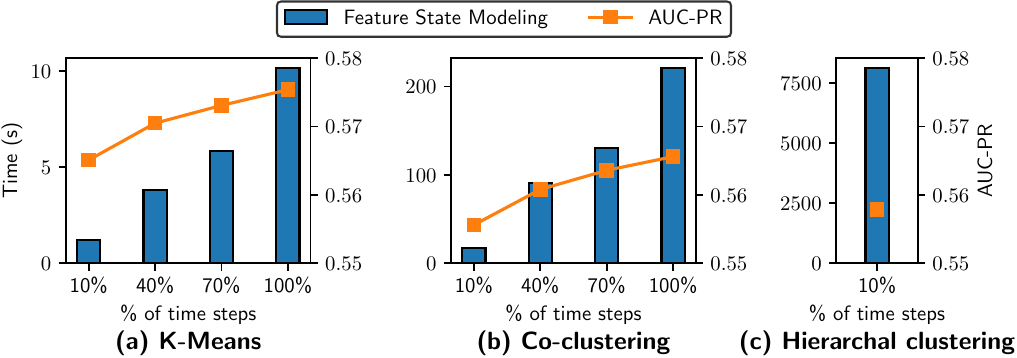}
\vspace{-5mm}
\caption{
Efficiency and performance analysis of different clustering techniques for feature state modeling in \CDM.
}
\vspace{-6mm}
\label{fig:cluster-compare}
\end{figure}

\subsection{Clustering algorithm comparison}
To justify the adoption of K-Means for feature state modeling within the \CDMfull, we conduct a comparative analysis against two alternative clustering techniques: co-clustering~\cite{dhillon2001co} and hierarchical clustering~\cite{johnson1967hierarchical},
which both group data into predefined numbers of clusters. To mitigate the issue of time complexity in clustering, we sample patient representations at varying sampling ratios of time step. 
As illustrated in Figure~\ref{fig:cluster-compare},
co-clustering not only incurs greater time consumption than K-Means but also yields inferior performance in terms of AUC-PR. Hierarchical clustering, meanwhile, demonstrates prohibitive time consumption when modeling just 10\% of time steps and suffers from memory exhaustion issues when handling larger volumes of data.
Further, both co-clustering and hierarchical clustering require extra computations to determine cluster centroids for evaluating new patients, while K-Means can streamline this process using learned centroids. K-Means also achieves better efficiency and effectiveness which confirms the its usage in our \CDMfull.

\section{Discussions} \label{sec:discussion}

\textbf{Selection of $k$ and $n$.}
In our experimental setup, the $k$ and $n$ are fixed across all features determined by prediction performance.
However, incorporating auxiliary information could improve their selections.
For instance, in feature state modeling, the selection of $k$ can be improved by considering feature characteristics such as missing rates and value ranges. For $n$ in subsequent cohort pattern modeling, employing thresholds on $\bm{\alpha}$ shows promise for automatically selecting $n$.
Hence, adapting these parameters based on feature characteristics and interactions may improve final performance.

\noindent
\textbf{Time complexity.} 
As analyzed in Section~\ref{subsec:complexity}, 
utilizing more patients, features, and time steps indeed facilitates the discovery of more meaningful cohorts and improves prediction. However, this concurrently introduces a higher complexity and extends the time required for identifying, learning, and exploiting cohorts. To mitigate the computational burden, we could consider implementing advanced cohort filters and iterative cohort update strategies to shorten cohort learning time.

\noindent
\textbf{Cohort validation and knowledge exploitation.} 
As shown in Section~\ref{sec:exp}, discovered cohorts can significantly improve the prediction performance and provide medically meaningful insights, often aligning with previous medical studies.
However, we acknowledge that not all high-performing cohorts demonstrate clear medical significance, necessitating further validation by clinicians. 

\bibliographystyle{ACM-Reference-Format}
\bibliography{sample-base}

\end{document}